\documentclass{article}
\usepackage{spconf,amsmath,graphicx}
\usepackage{subfigure}
\usepackage{cite}
\usepackage{hyperref}
\usepackage{breakurl}
\usepackage{color}

\title{Security of Facial Forensics Models Against Adversarial Attacks}
\name{Rong Huang\textsuperscript{1}, Fuming Fang\textsuperscript{2}, Huy H. Nguyen\textsuperscript{3}, Junichi Yamagishi\textsuperscript{2,3}, Isao Echizen\textsuperscript{2,3}\thanks{This work was partially supported by a JST CREST Grant (JPMJCR18A6, VoicePersonae project), Japan, and by MEXT KAKENHI Grants (16H06302, 17H04687, 18H04120, 18H04112, 18KT0051), Japan.}}
\address{\textit{\textsuperscript{1}College of Information Science and Technology, Donghua University, Shanghai, China} \\
\textit{\textsuperscript{2}National Institute of Informatics, Tokyo, Japan}\\
\textit{\textsuperscript{3}SOKENDAI (The Graduate University for Advanced Studies), Kanagawa, Japan}}
\begin{document}
\ninept
%

\maketitle
\begin{abstract}
Deep neural networks (DNNs) have been used in digital forensics to identify fake facial images. We investigated several DNN-based forgery forensics models (FFMs) to examine whether they are secure \linebreak against adversarial attacks. We experimentally demonstrated the existence of individual adversarial perturbations (IAPs) and universal adversarial perturbations (UAPs) that can lead a well-performed FFM to misbehave. Based on iterative procedure, gradient information is used to generate two kinds of IAPs that can be used to fabricate classification and segmentation outputs. In contrast, UAPs are generated on the basis of over-firing. We designed a new objective function that encourages neurons to over-fire, which makes UAP generation feasible even without using training data. Experiments demonstrated the transferability of UAPs across unseen datasets and unseen FFMs. Moreover, we conducted subjective assessment for imperceptibility of the adversarial perturbations, revealing that the crafted UAPs are visually negligible. These findings provide a baseline for evaluating the adversarial security of FFMs.
\end{abstract}
\vspace{-0.1cm}
\section{Introduction}
\vspace{-0.1cm}
\label{sec:intro}

Faces appearing in digital media are salient information that directly reflects personal identity and scene content. Facial manipulation programs, like DeepFakes \cite{df}, Face2Face and FaceSwap \cite{fs}, can be used to automatically edit facial identities and expressions so as to alter the semantic content of the digital media. 
To deal with this security threat, many countermeasures \cite{lip,eye,head,iqm,color,warping,MesoNet,Mo,FT,BTAS} have been proposed. Inspired by presentation attack detection, researchers are attempting to use liveness clues like lip-syncing \cite{lip}, eye blinking \cite{eye}, and inconsistency in head pose \cite{head} to identify fake facial videos. Another approach is to use inconsistent statistical characteristics including IQM (image quality metrics) descriptor \cite{iqm}, color disparity \cite{color}, and warping artifacts \cite{warping} as telltale indicators for detecting forgeries. DNNs have also been used to expose fake facial media \linebreak \cite{Mo,MesoNet}. To enhance transferability, Cozzolino {\em et al.} \cite{FT} designed a selection-based domain adaptation mechanism over the latent \linebreak space of an autoencoder. A variation of this approach \cite{BTAS}, using a multi-task autoencoder, was proposed for simultaneously achieving the forensic goals of forgery identification and localization. The reported DNN-based FFMs \cite{Mo,MesoNet,FT,BTAS} have exhibited outstanding performance and shown great potential in security-sensitive intelligent systems, such as fake news identification and electronic evidence investigation.

\par However, the security of these DNN-based FFMs \cite{Mo,MesoNet,FT,BTAS} is problematic since an attacker may carefully fabricate a fake facial image or video to circumvent the intelligent systems for getting higher but illegal privilege. Recent works \cite{survey1,survey2} have revealed that adding a tailor-made tiny noise to the bona fide input can easily induce a well-performed DNN to produce misleading output. Previous works on adversarial attacks mainly focused on tasks involving face recognition \cite{face_attack}, object detection \cite{obj_attack}, semantic segmentation \cite{ss_attack}, natural language processing \cite{nlp_attack}, and malware detection \cite{md_attack}. There has been a lack of work on whether FFMs are secure against adversarial attacks. We aim to fill this gap, in this paper, by applying gradient-based adversarial attacks (white box) to Nguyen's FFM \cite{BTAS} and \linebreak evaluating the transferability of adversarial perturbations (black box) across unseen datasets and unseen FFMs \cite{MesoNet,FT}.
\par Specifically, our contributions are as follows: {\bf{1.}} We show the existence of IAPs, which can be used to fabricate classification or segmentation outputs. {\bf{2.}} We show the existence of image-agnostic UAPs for the FFM \cite{BTAS}. We present a newly designed objective function that enables latent neurons to be over-fired without the need for training data. {\bf{3.}} We present experimental results demonstrating that UAPs have transferability across unseen datasets and unseen FFMs \cite{MesoNet,FT}. {\bf{4.}} Following an international standard \cite{ITU}, we conduct subjective assessment for imperceptibility of adversarial perturbations. This is a necessary step oriented towards practical applications of the adversarial attack.
\par The experimental results presented in Section \ref{sec:main} support the above claims and thereby demonstrate that existing FFMs \cite{MesoNet,FT,BTAS} are vulnerable to adversarial attacks. This work provides a warning that future digital forensics research should take into account the need to integrate an adversarial defense mechanism into FFMs.

\vspace{-0.1cm}
\section{Background}
\vspace{-0.1cm}
\label{sec:GB}
In this section, we give a brief introduction of the target facial FFMs \cite{MesoNet,FT,BTAS}, and then introduce fundamental concepts and impressive achievements in the field of adversarial attack. 

\vspace{-0.08cm}
\subsection{Target FFMs}
\vspace{-0.08cm}
\label{ssec:TF}
In \cite{MesoNet}, Afchar {\em et al.} constructed a lightweight DNN with four convolutional layers and two fully connected layers, and furthermore proposed an advanced network structure called `MesoInception-4'. Inception modules consisting of vanilla convolutions, dilated convolutions and skip-connections ($1\times1$ convolutions) were introduced for adaptively probing low-level image textures.
\par The ForensicTransfer method \cite{FT} is based on an autoencoder architecture. For a given image $x_i$, its residual map that only contains high-frequency details is regarded as the input. Neurons in the latent layer are divided equally into two disjoint portions: zero zone $h_{i,0}$ and one zone $h_{i,1}$. The zone-wise activation energy is calculated by $a_{i,c}=\|h_{i,c}\|_1/K_c$, where $K_c$ counts the number of neurons in the $c^{\text{th}}$ zone for $c\in\{0,1\}$. The activation loss for an image $x_i$ is defined as $\mathcal{L}_{\text{act}}(x_i,l_i) =  | a_{i,1}-l_i | + | a_{i,0}-(1-l_i) |$, where $l_i$ represents the category label, which takes the value 0 for `fake' and 1 for `real'. The activation loss encourages the zone $h_{i,c}$  that $c=l_i$ to activate ($a_{i,c}>0$) while the other zone $h_{i,1-c}$ remains silent ($a_{i,1-c}=0$).
During the training phase, the selection module compulsively zeros out the off-class zone $h_{i,1-c}$, which forces the decoder to complete its reconstruction task using only $h_{i,c}$. During the testing phase, an image $x_i$ is classified as $c$ if $a_{i,c}>a_{i,1-c}$. This encoder-selection-decoder framework forces the latent space to learn discriminative representations and thus enhances the transferability across multiple domains of image manipulations.

\par In \cite{BTAS}, Nguyen {\em et al.} proposed a Y-shaped network that extends the network of Cozzolino {\em et al.} \cite{FT} in three ways. First, the residual module is abandoned so that a variety of features can be adaptively probed by convolutional layers without constraints. Second, the autoencoder architecture is made deeper (with up to 18 convolutional layers). Third, the decoder is designed as a two-pronged structure, which enables three tasks (classification, segmentation, and reconstruction) to share the information they learn and thereby enhance overall performance. The segmentation branch is responsible for locating forged regions. Cross-entropy loss $\mathcal{L}_{\text{seg}}(x_i,m_i)$ is used to evaluate the segmentation quality, where $m_i$ represents the ground-truth mask.

\par In this paper, the above three representative FFMs \cite{MesoNet,FT,BTAS} are considered as the target models for examining the existence and effectiveness of adversarial perturbations. FaceForensics++ database \cite{data} was used to train or fine-tune the three FFMs. It consists of three datasets, `DeepFakes', `Face2Face' and  `FaceSwap'. Each dataset contains 1000 real facial videos and 1000 manipulated ones. This database also provides ground-truth masks that highlight the manipulated regions in a pixel-wise manner. Data preparation followed the procedure previously used in \cite{FT,BTAS}, so there were 2800 images in each test set that would be used to evaluate the performance of adversarial attacks. 

\vspace{-0.08cm}
\subsection{Adversarial Attacks}
\vspace{-0.08cm}
Given an image $x_i$ and a target model $f$, an adversarial attack aims to find a tiny perturbation, denoted by $\xi_i$, so that the corresponding adversarial image $x^{\text{adv}}_i=x_i+\xi_i$ can change the output of $f$, i.e., $f(x_i)\neq f(x^{\text{adv}}_i)$. During the course of seeking $\xi_i$, its magnitude, which is usually measured by the $p$-norm distance, is bounded by a small value to ensure its imperceptibility.

\par In the pioneering work \cite{BFGS}, Szegedy {\em et al.} generated adversarial perturbations by using a box-constrained L-BFGS. However, this method is relatively time-consuming. A suitable value of hyperparameter $c$ that modulates the balance between two terms in the objective function is unknown. This therefore requires an additional line-searching to adaptively update $c$.

\par Fast gradient sign method (FGSM) proposed by Goodfellow {\em et al.} \cite{FGSM} is an efficient one-time method without iterations. An adversarial image is generated by slightly adapting each pixel along the direction of the sign of gradient.
Basic iterative method (BIM) \cite{BIM} is an intuitive extension of FGSM. It iteratively generates an adversarial perturbation with a smaller step size, which enables the updating direction to be adjusted after each iteration. 

\par Deepfool \cite{Deepfool} was proposed for finding a minimal perturbation $\xi_i$ that could just push $x_i$ outside its own classification region. An iterative procedure is used to gradually estimate $\xi_i$ under the assumption that the linearity of $f$ holds around $x_i$ at each iteration. In \cite{UAP}, Moosavi-Dezfooli {\em et al.} leveraged the Deepfool method to craft an image-agnostic UAP $\Xi$ that can fool the target model $f$, i.e., $f(x_i)\neq f(x_i+\Xi)$, for a vast majority of $x_i$. 
The crafted UAP was empirically demonstrated to transfer well across multiple architectures including VGG, CaffeNet, GoogLeNet, and ResNet. 

\par The idea of Mopuri {\em et al.}'s work \cite{overfire} is to seek a UAP that can over-fire the neurons (maximizing the activation energies) at each layer. Such a UAP would contaminate the learned features and thereby lead the target model to misbehave. Particularly interesting is that the training phase for over-firing can be done in a data-free manner.

\par In this paper, the gradient information itself (rather than the signed version) was used for generating IAPs. Inspired by the idea of over-firing, we crafted UAPs without using training data. However, we found that, for autoencoder network \cite{BTAS}, it is difficult to achieve over-firing due to the small magnitude of the perturbation-only input. We solved this problem by designing a new objective function that strongly promotes over-firing of the latent neurons in the desired zone (see Section \ref{s2}).



			
			


\vspace{-0.1cm}
\section{Adversarial Perturbations For FFMs}
\vspace{-0.1cm}
\label{sec:main}
In this section, we mount the adversarial attacks against the FFM \cite{BTAS} for different falsification purposes. First, we craft two kinds \linebreak of IAPs for each image, one for fabricating classification output and one for fabricating segmentation output. Next, we craft UAPs in accordance with the idea of over-firing and evaluate their transferability across unseen datasets and unseen FFMs \cite{MesoNet,FT}. Finally, we describe the settings for the subjective assessment and summarize the insights gained from the assessment into the imperceptibility of adversarial perturbations.

\vspace{-0.08cm}
\subsection{Individual adversarial attacks}
\vspace{-0.08cm}
\label{s1}
Given an image $x_i$, we iteratively generate a tailor-made IAP that can flip the original classification output. The iterative procedure can be written as:
\begin{equation}
\xi_i^{(n+1)}=\text{Clip}_{\epsilon}\big\{\xi_i^{(n)}-\alpha(\nabla_{x_i} \mathcal{L}_{\text{act}}(x_i+\xi_i^{(n)},l^{\text{adv}}_i)) \big\},
\label{g_IAP}
\end{equation}
where $\xi_i^{(n)}$ denotes the adversarial perturbation at the $n^{\text{th}}$ iteration while $\alpha$ represents the step size. The iterative procedure starts with $\xi_i^{(0)}=\mathbf{0}$. At each iteration, the operator $\text{Clip}_{\epsilon}(\cdot)$ restricts the perturbation's magnitude within $[-\epsilon,\epsilon]$ to ensure imperceptibility. Adversarial label $l^{\text{adv}}_i$ satisfying $l^{\text{adv}}_i \neq l_i$ enables the fabricated output to be controlled. For a binary classification problem (`real' or `fake'), it is easy to prepare the adversarial label: $l^{\text{adv}}_i=1-l_i$. The sign operator previously used in FGSM and BIM is excluded from our iterative procedure since we consider that not only the direction but also the length of the gradient is informative in generating IAPs. Note that only the activation loss is used in Eq.(\ref{g_IAP}) because the classification is performed in accordance with the zone-wise activation energy. We set $\alpha=1.0$ and $\epsilon=2.5$ throughout our experiments.

\par
\begin{figure}[t]
    \centering

    \subfigure{\includegraphics[height=0.4cm]{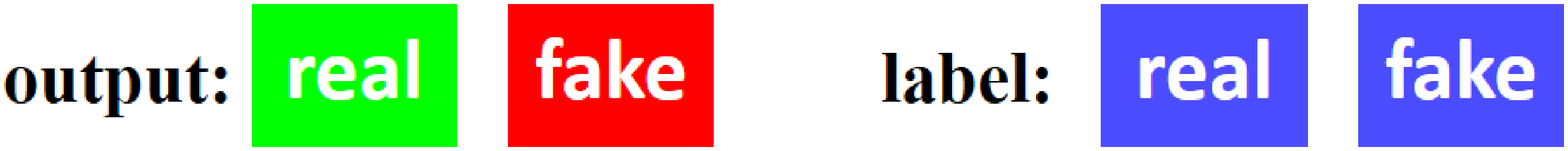}}\addtocounter{subfigure}{-1}
    \vspace{-0.3cm}

    \subfigure{\includegraphics[height=1.65cm]{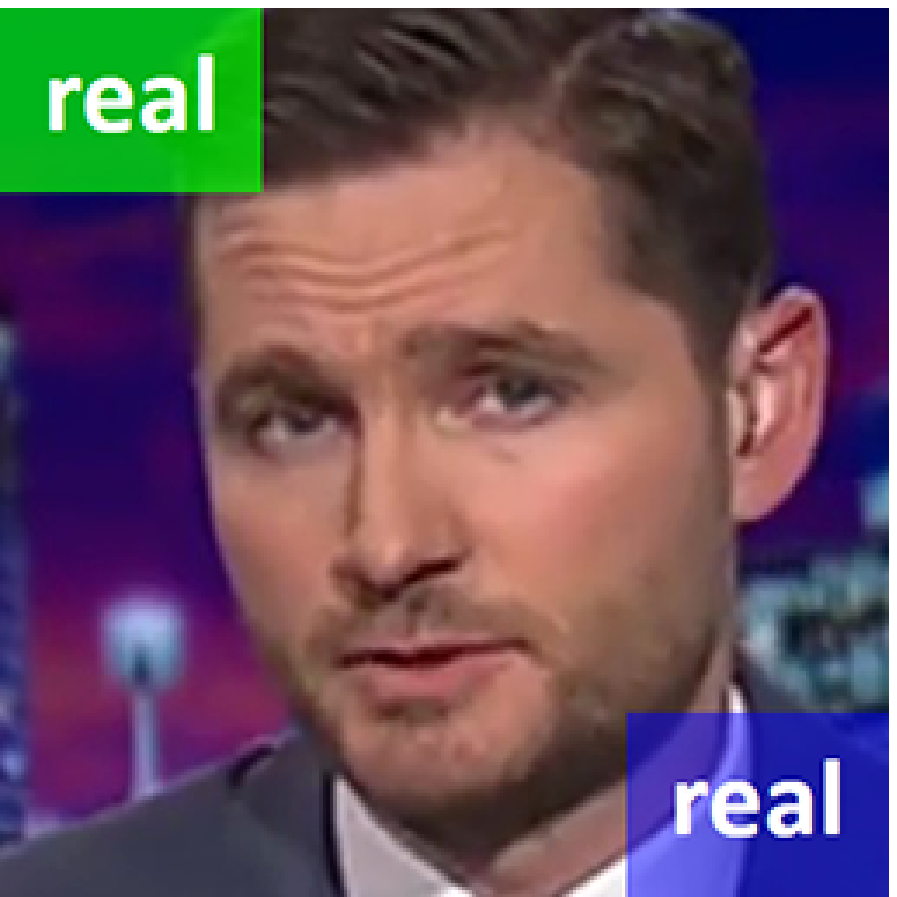}}
    \hspace{-0.12cm}
    \subfigure{\includegraphics[height=1.65cm]{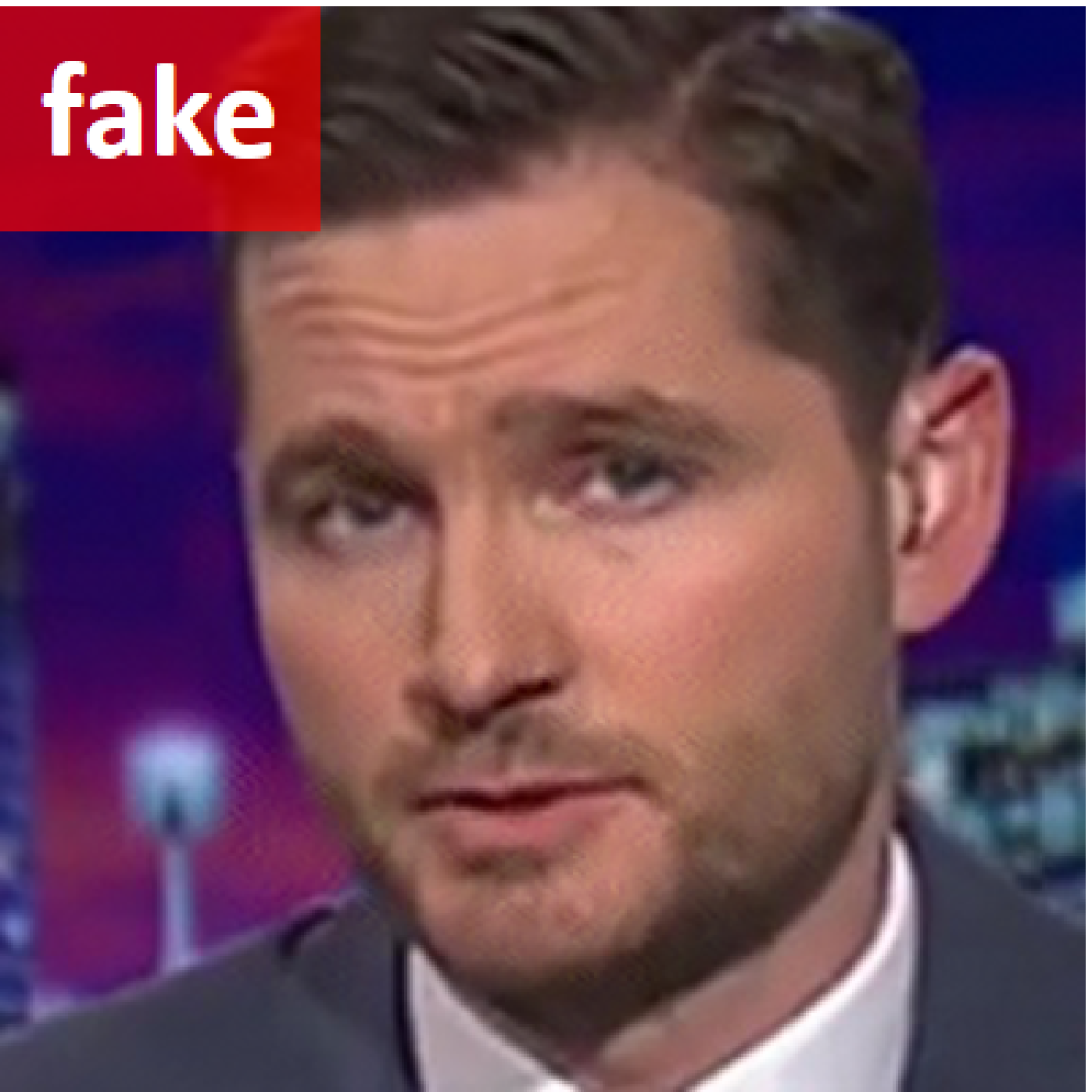}}
    \hspace{-0.12cm}
    \subfigure{\includegraphics[height=1.65cm]{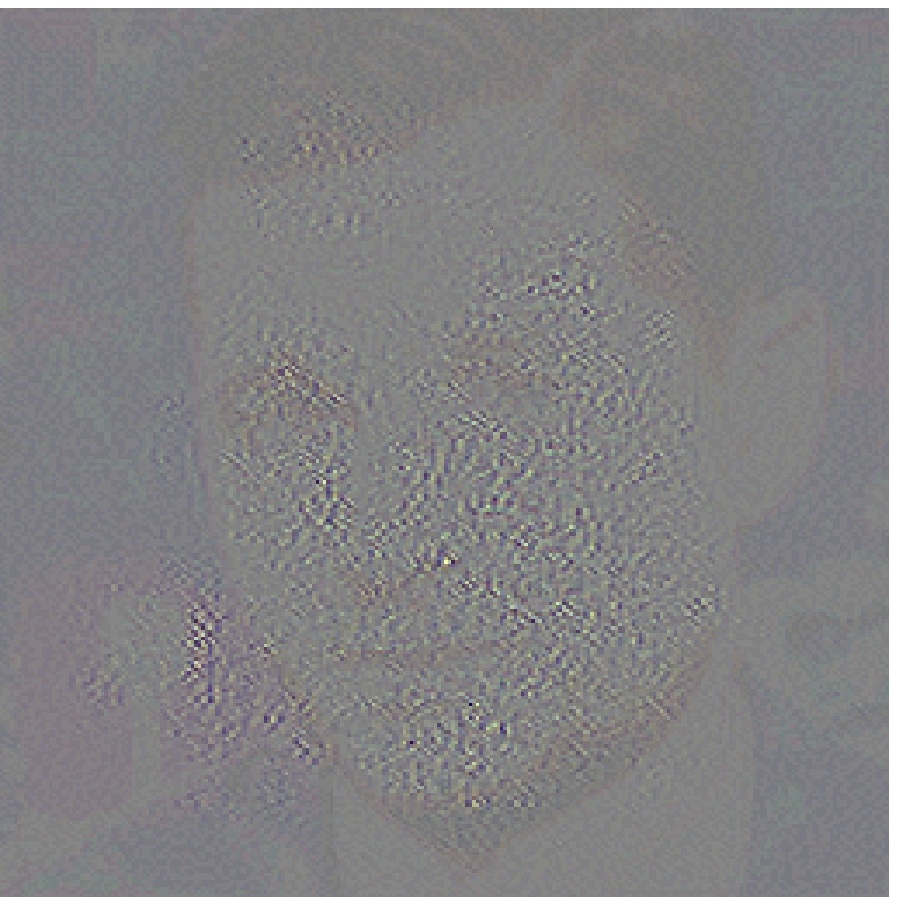}}
    \hspace{-0.12cm}
    \subfigure{\includegraphics[height=1.65cm]{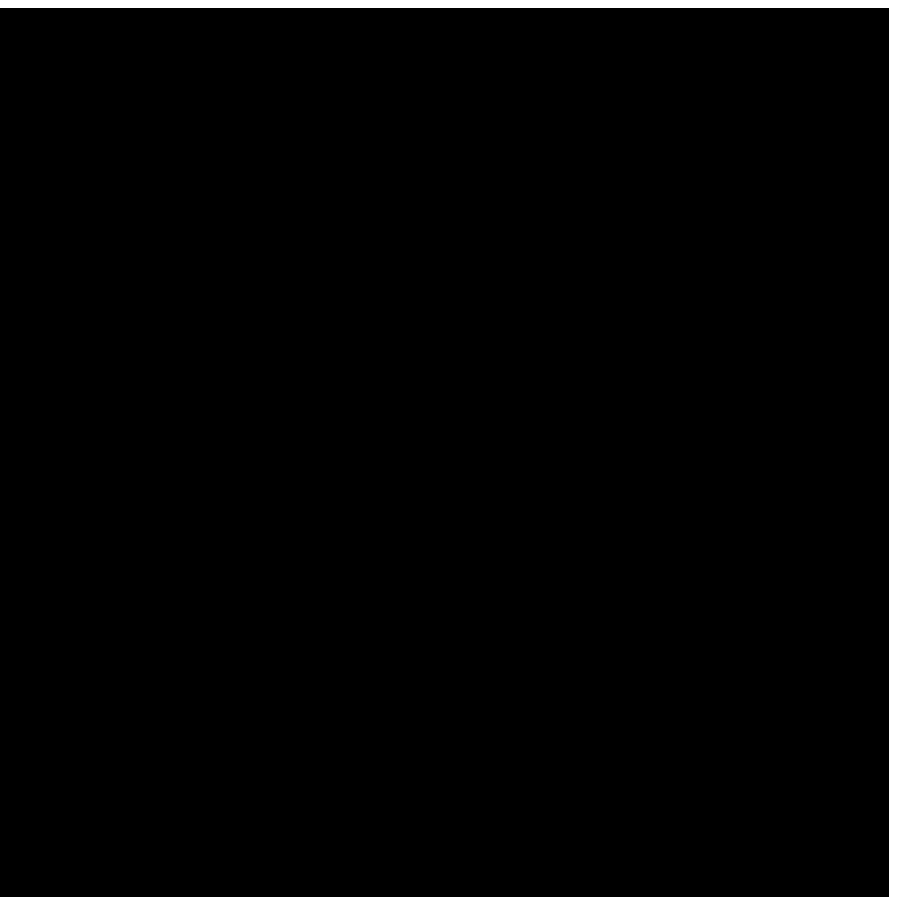}}
    \hspace{-0.12cm}
    \subfigure{\includegraphics[height=1.65cm]{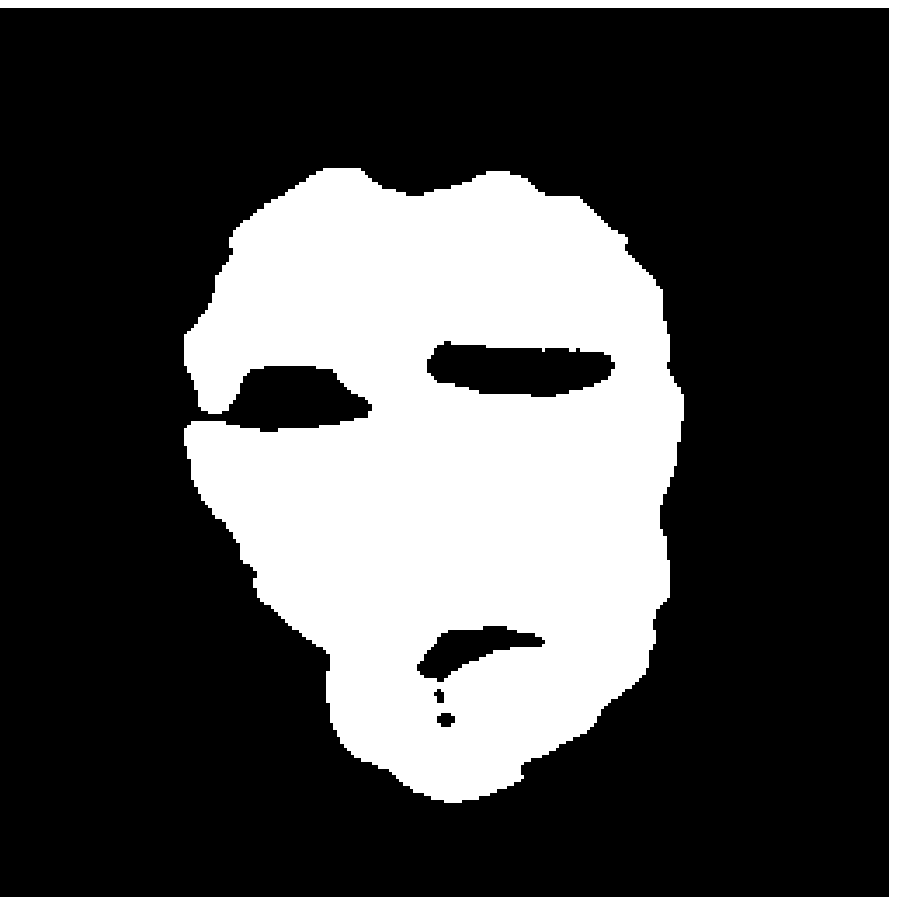}}\addtocounter{subfigure}{-5}



    \vspace{-0.3cm}

    \subfigure[]{\includegraphics[height=1.65cm]{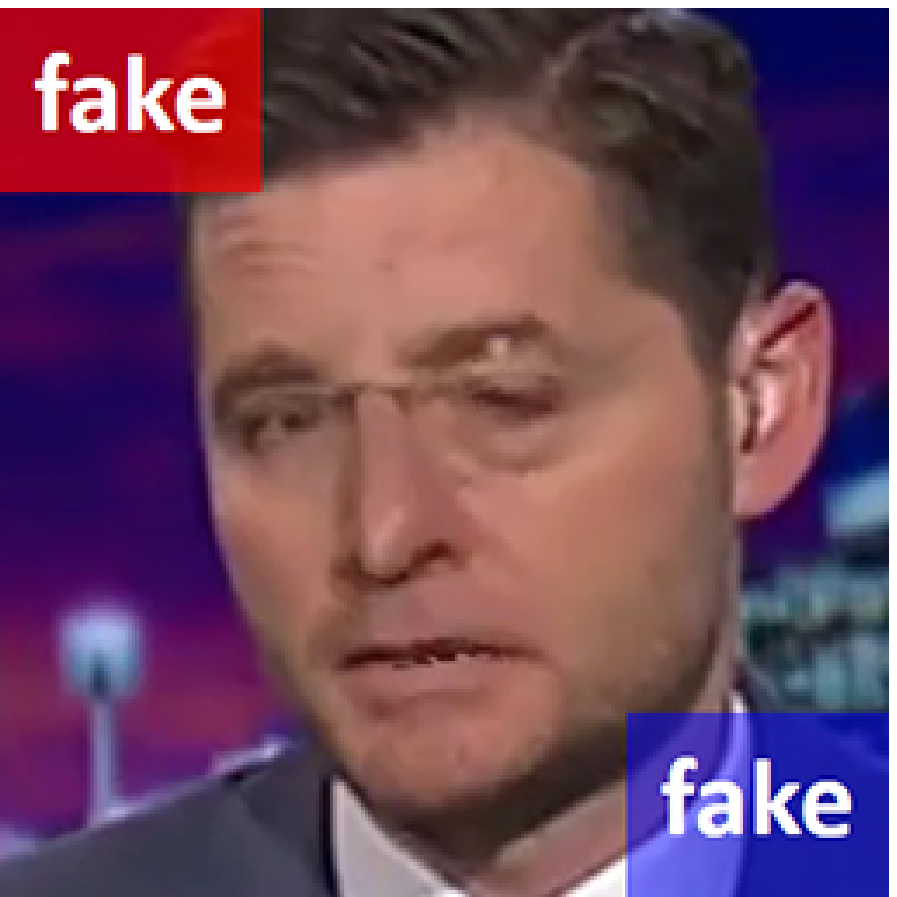}}
    \hspace{-0.12cm}
    \subfigure[]{\includegraphics[height=1.65cm]{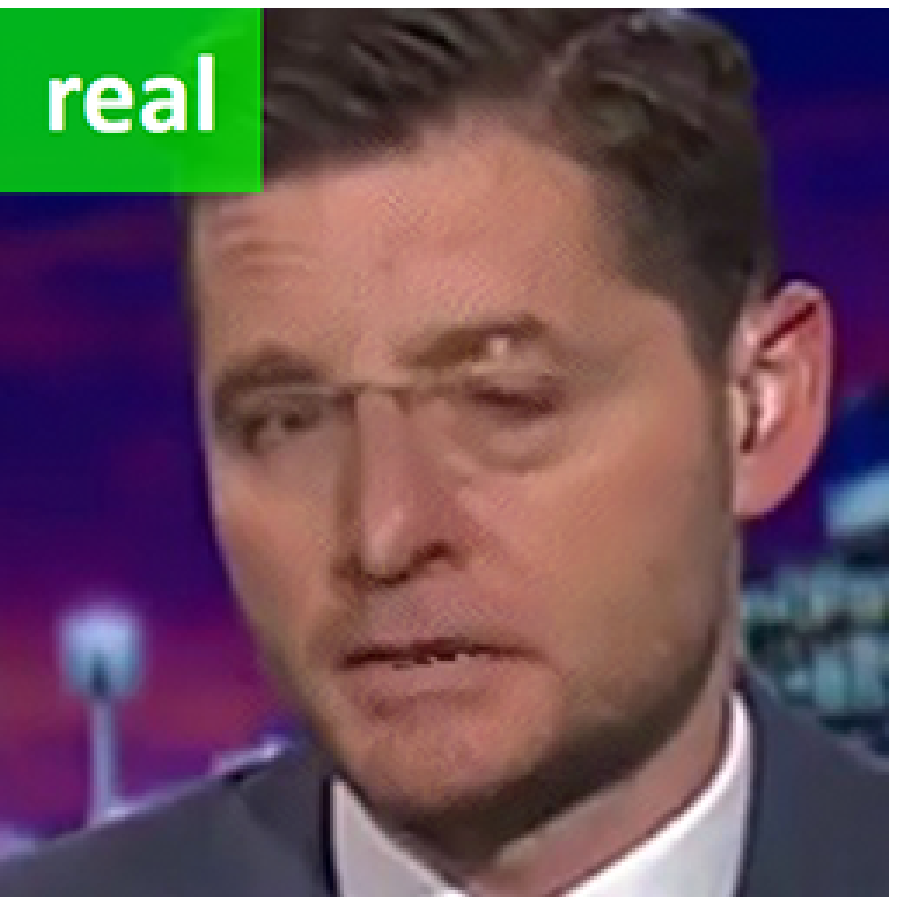}}
    \hspace{-0.12cm}
    \subfigure[]{\includegraphics[height=1.65cm]{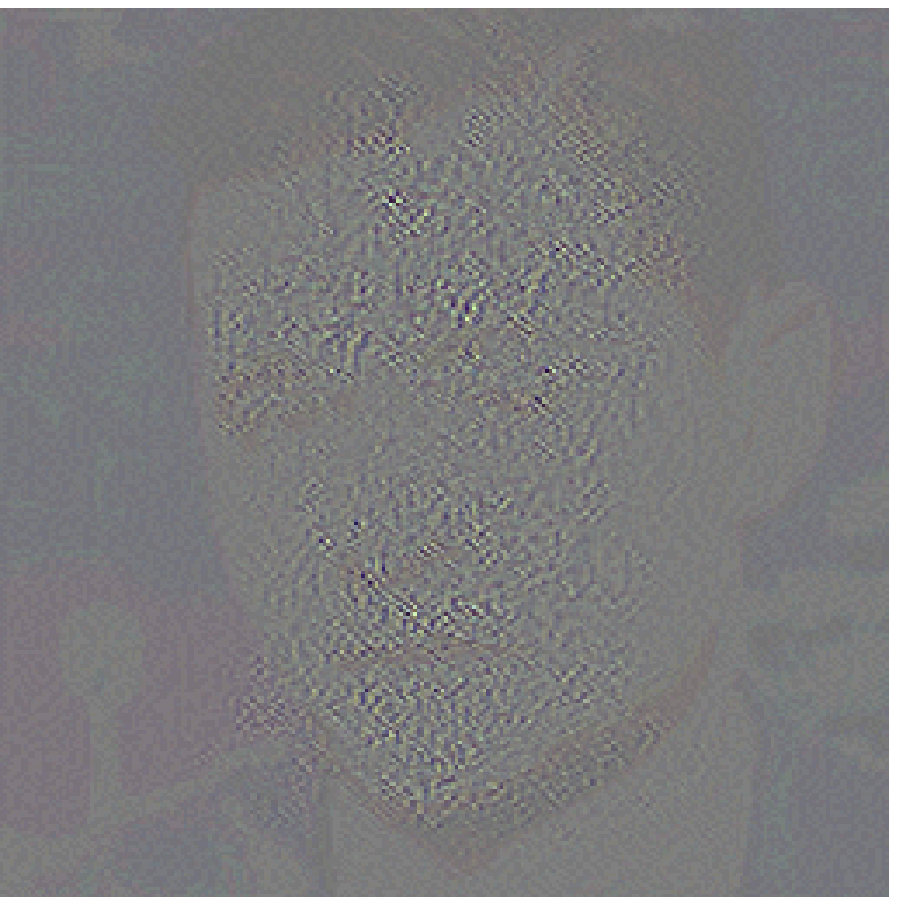}}
    \hspace{-0.12cm}
    \subfigure[]{\includegraphics[height=1.65cm]{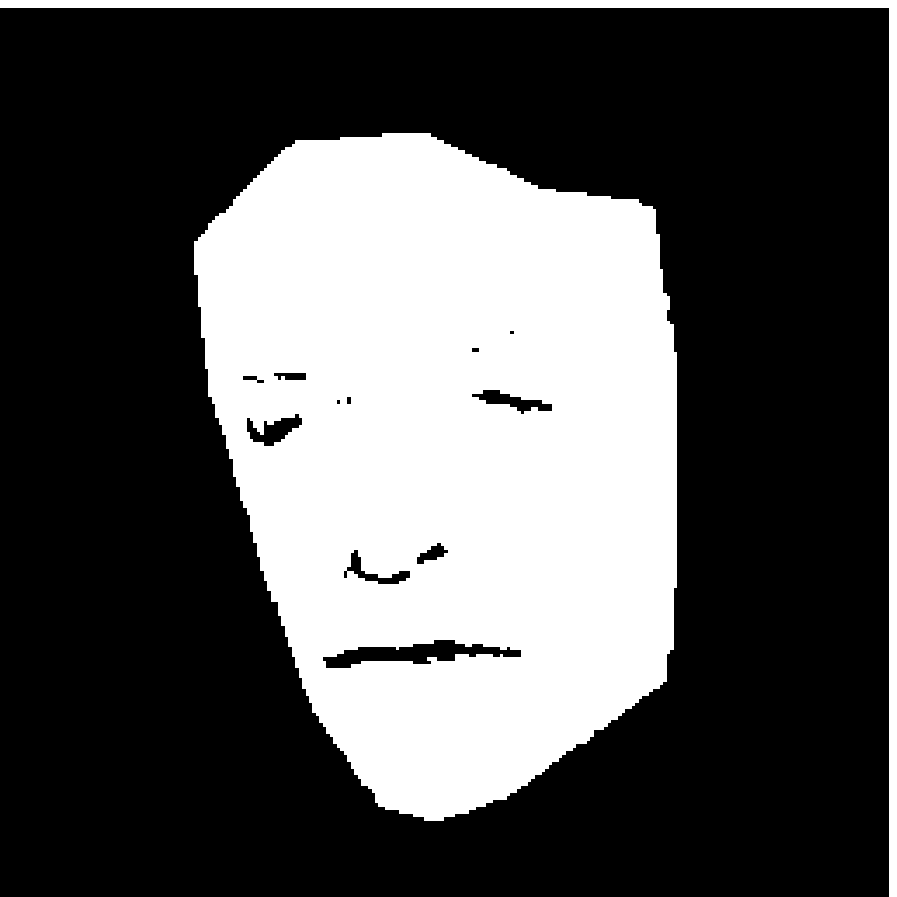}}
    \hspace{-0.12cm}
    \subfigure[]{\includegraphics[height=1.65cm]{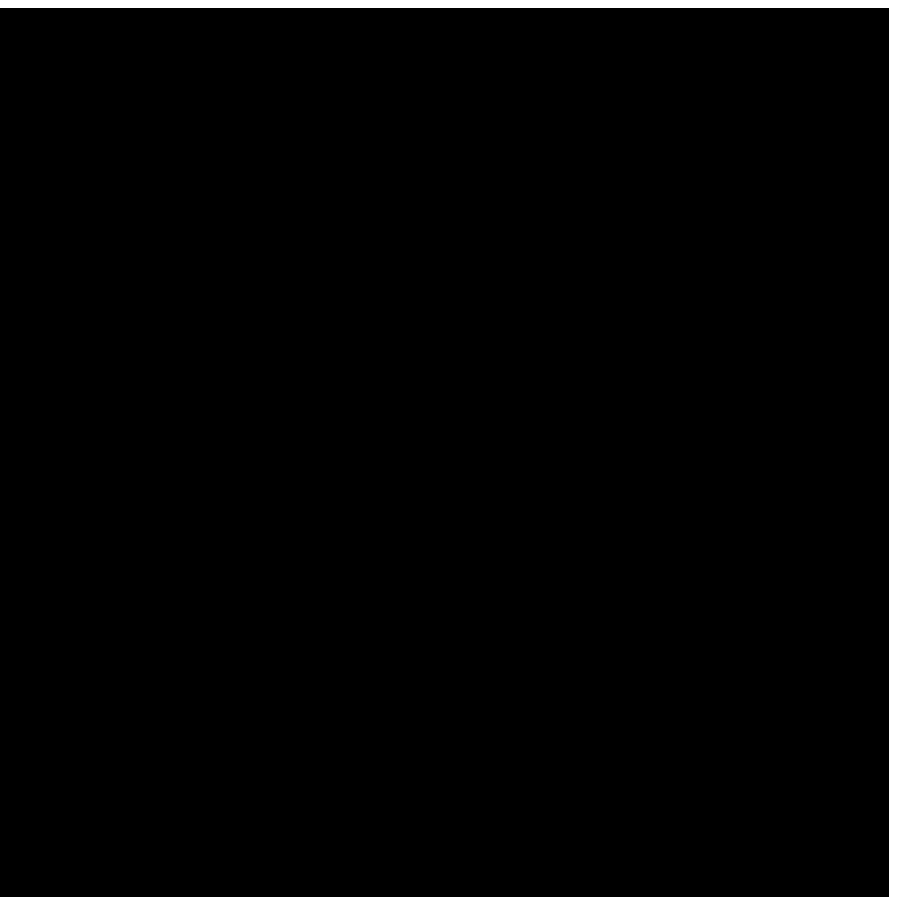}}\addtocounter{subfigure}{-5}



    \vspace{-0.4cm}

    \caption{Resulting examples (selected from: FaceSwap \& 20 iters). (a) Original images. (b) Adversarial images. (c) Adversarial perturbations (min-max scaled for display). (d) Ground-truth masks. (e) Segmentation outputs.}
    \label{fig:IAP_c}
\vspace{-0.5cm}
\end{figure}

 Table \ref{tab_IAP_c} shows that adding IAPs drastically reduced classification accuracies. The small RMSE (root mean squared error) scores, however, suggest perceptual similarities between $x_i$ and $x^{\text{adv}}_i$. More importantly, 20 iterations for each image were sufficient to destory the well-performed FFM \cite{BTAS} over the three datasets. Furthermore, the Intersection-over-Union (IoU) scores were dramatically suppressed at the same time although the segmentation loss $\mathcal{L}_{\text{seg}}(x_i,m_i)$ was absent from the iterative procedure. This is because only activated latent features $h_{i,c}$ were selected for the following decoder, so the classification output greatly affected the segmentation branch. This phenomenon is reflected in the two resulting examples shown in Fig.\ref{fig:IAP_c}, in which the segmentation outputs, i.e., the column (e), \linebreak agree with the fabricated classification outputs.

\begin{table}[t]
	\begin{center}

        \caption{Performance of IAPs (fabricating classification outputs).}
        \vspace{-0.00cm}
        \begin{tabular}{c}
			
			\scalebox{0.23}[0.23]{\includegraphics{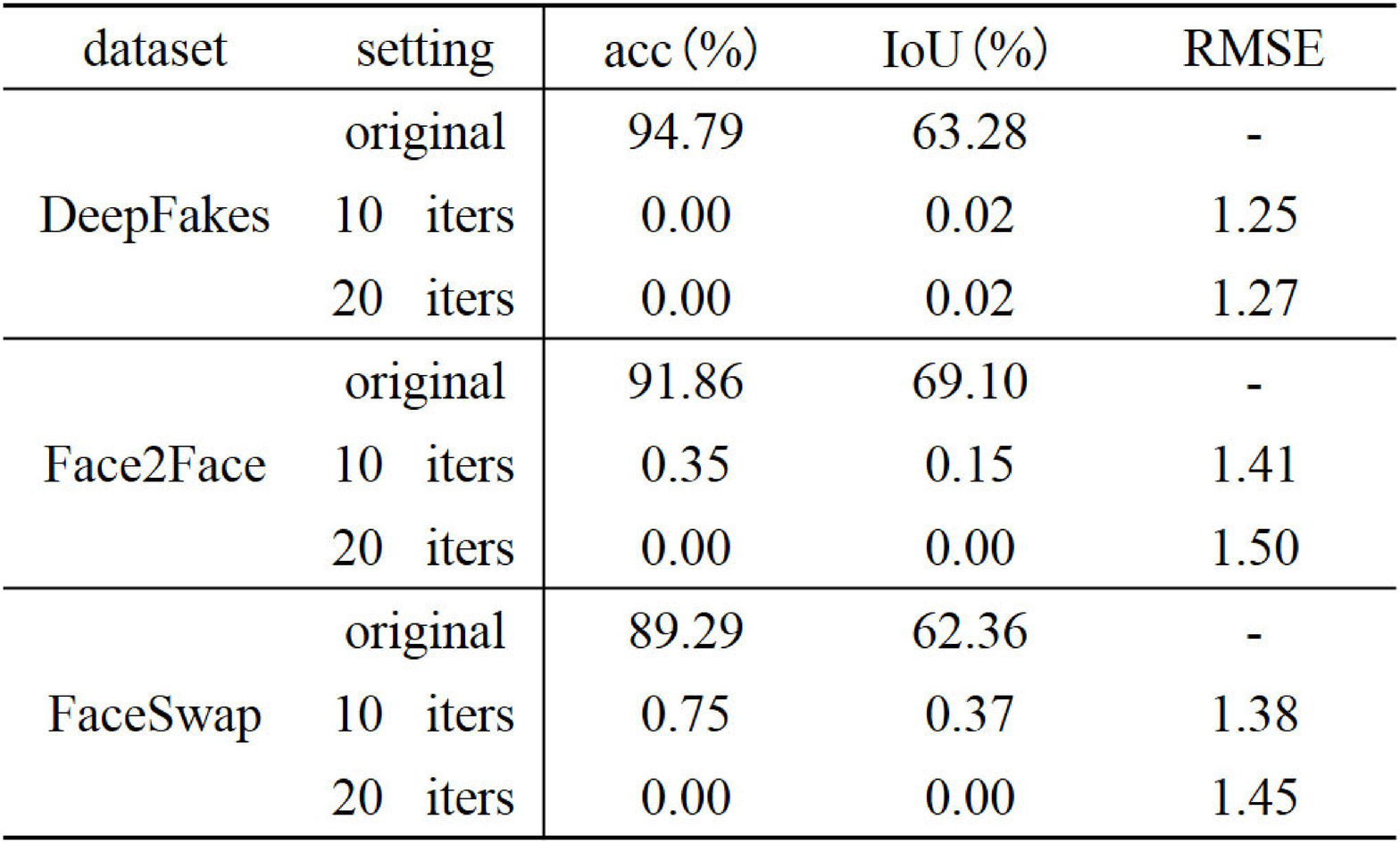}}
			
		\end{tabular}

		\label{tab_IAP_c}
	\end{center}
\vspace{-0.7cm}
\end{table}

\par
\begin{figure}[t]
    \centering

    \subfigure{\includegraphics[height=1.65cm]{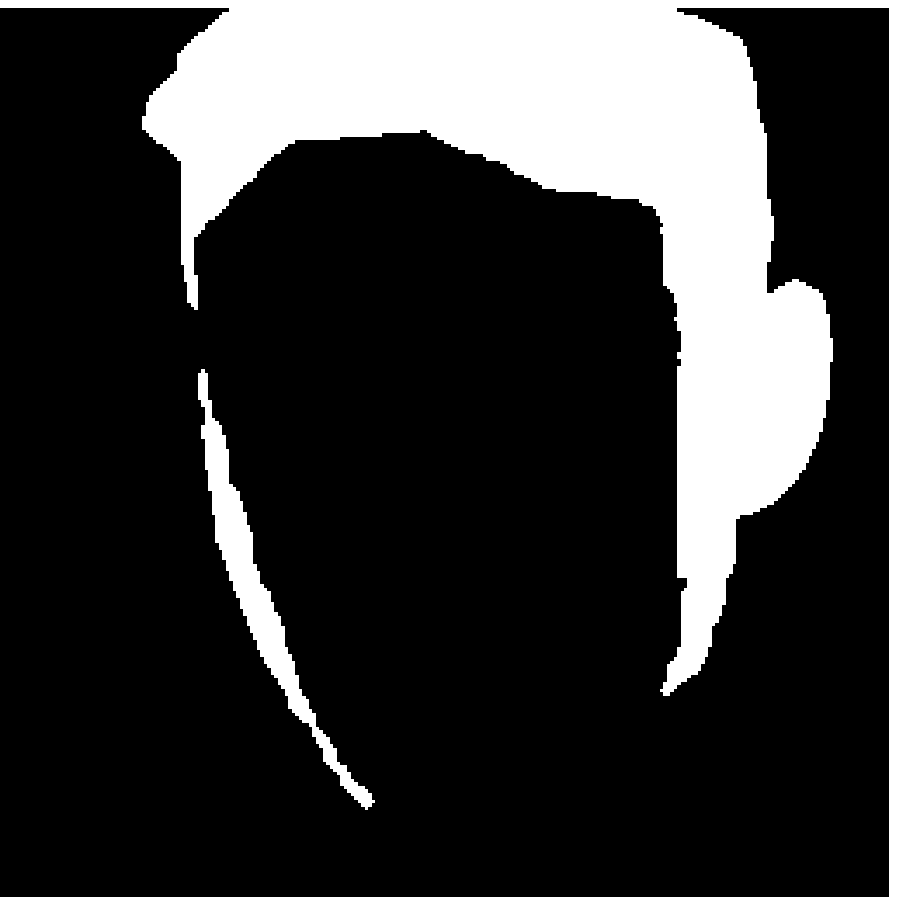}}
    \hspace{-0.08cm}
    \subfigure{\includegraphics[height=1.65cm]{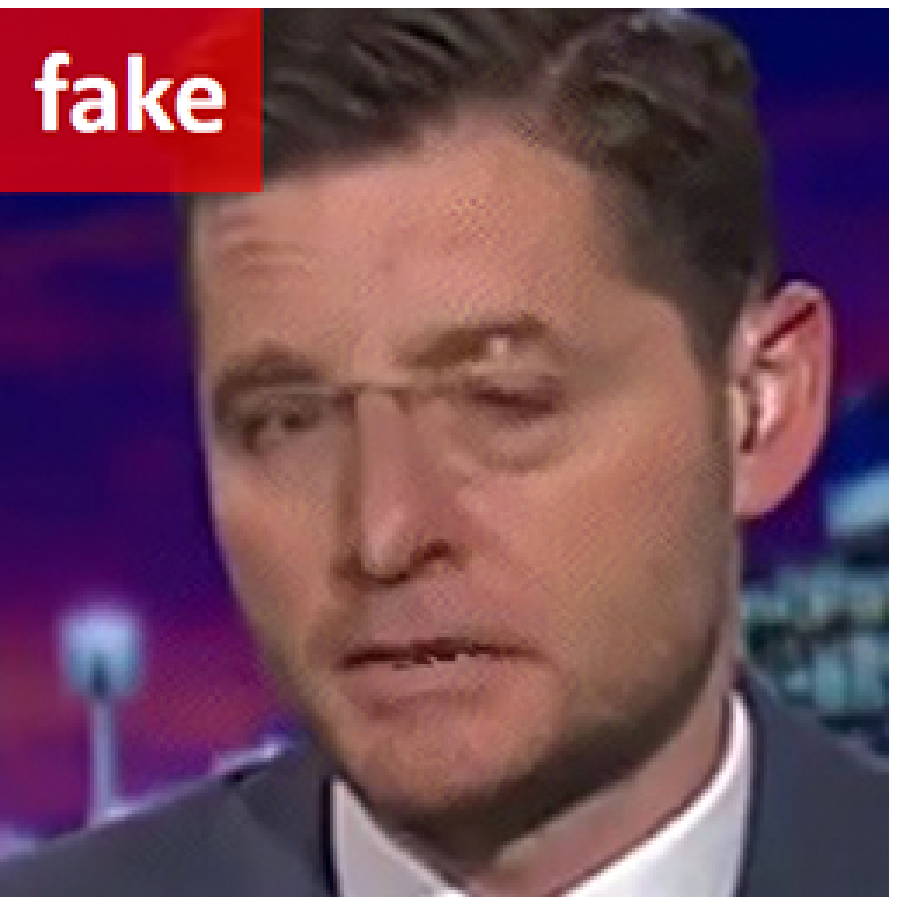}}
    \hspace{-0.08cm}
    \subfigure{\includegraphics[height=1.65cm]{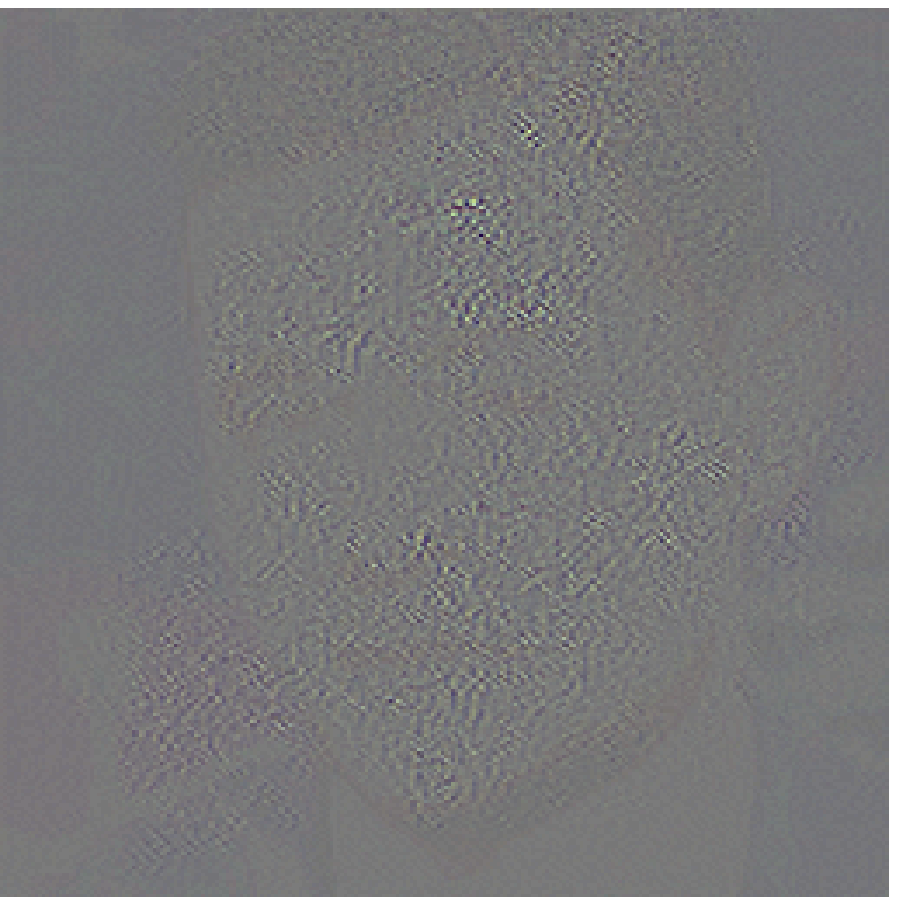}}
    \hspace{-0.08cm}
    \subfigure{\includegraphics[height=1.65cm]{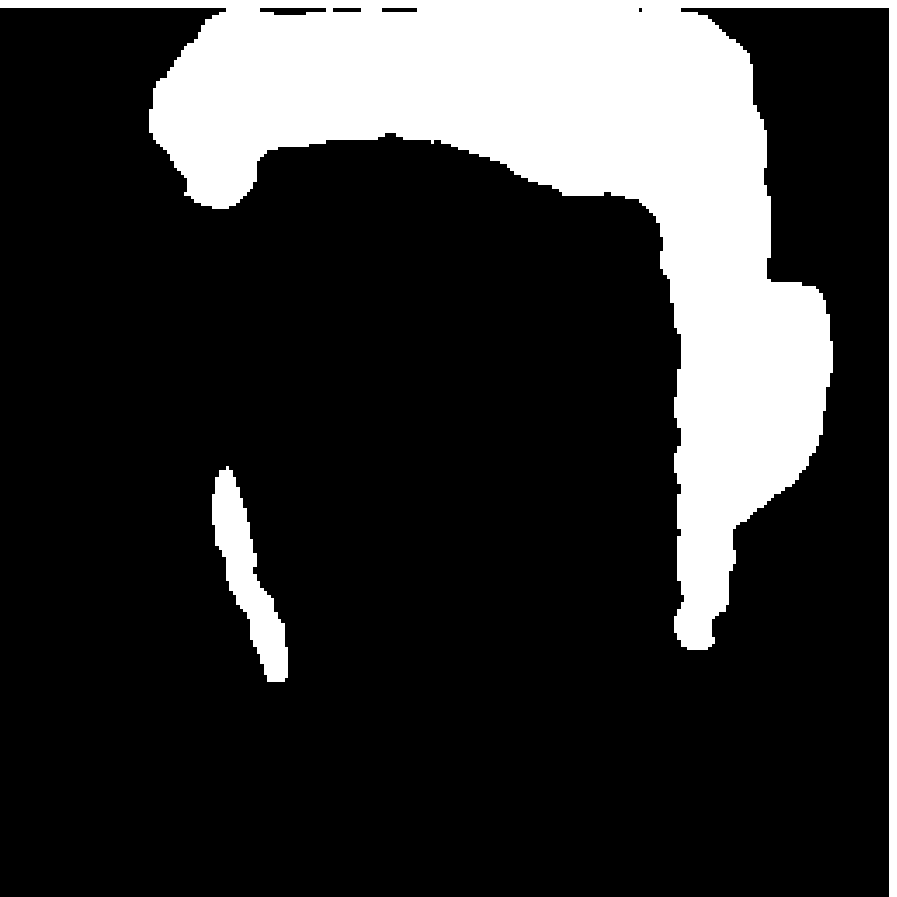}}\addtocounter{subfigure}{-4}



    \vspace{-0.3cm}

    \subfigure[]{\includegraphics[height=1.65cm]{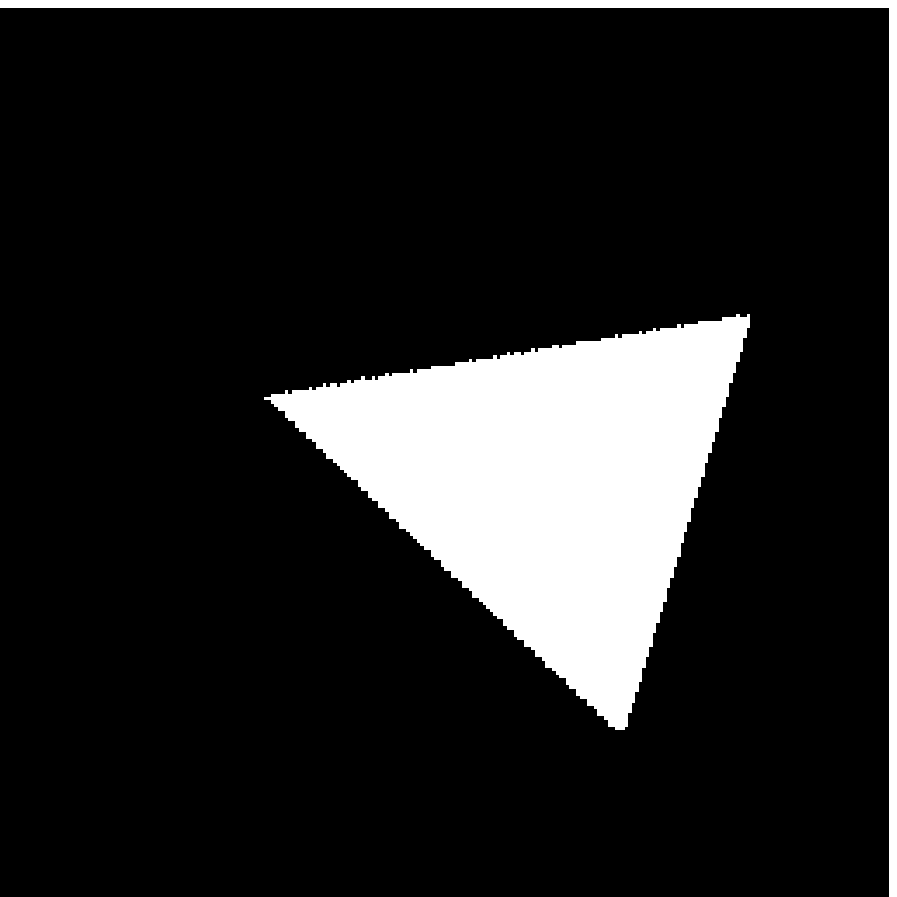}}
    \hspace{-0.08cm}
    \subfigure[]{\includegraphics[height=1.65cm]{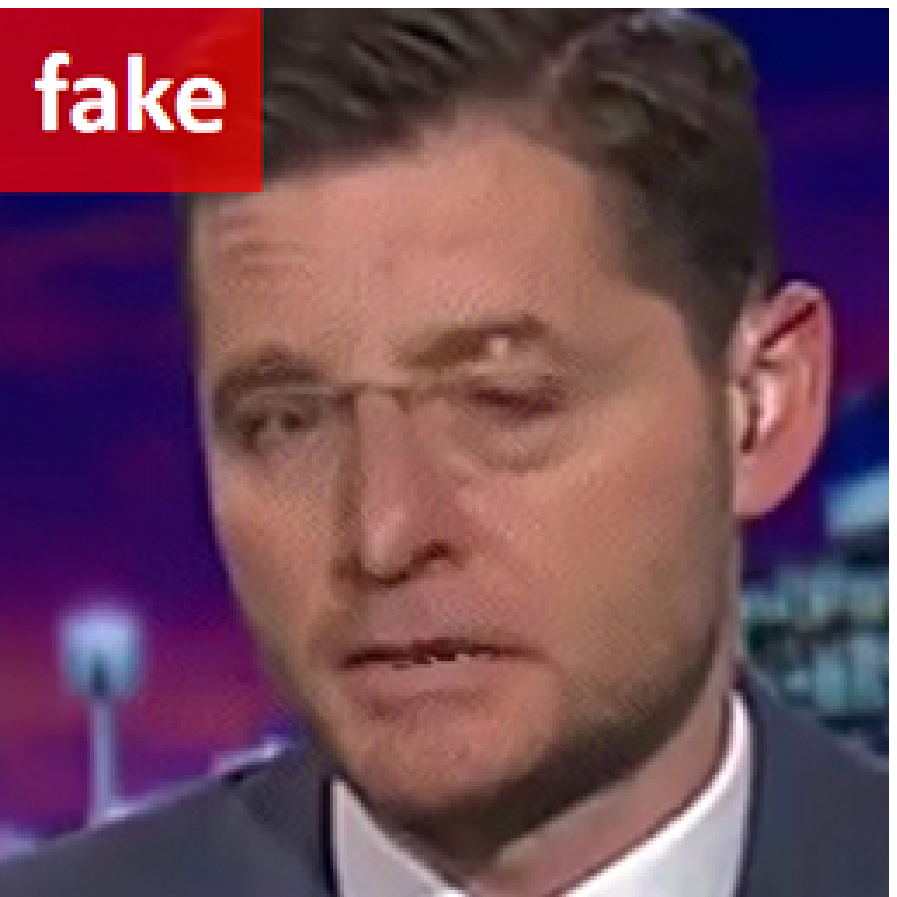}}
    \hspace{-0.08cm}
    \subfigure[]{\includegraphics[height=1.65cm]{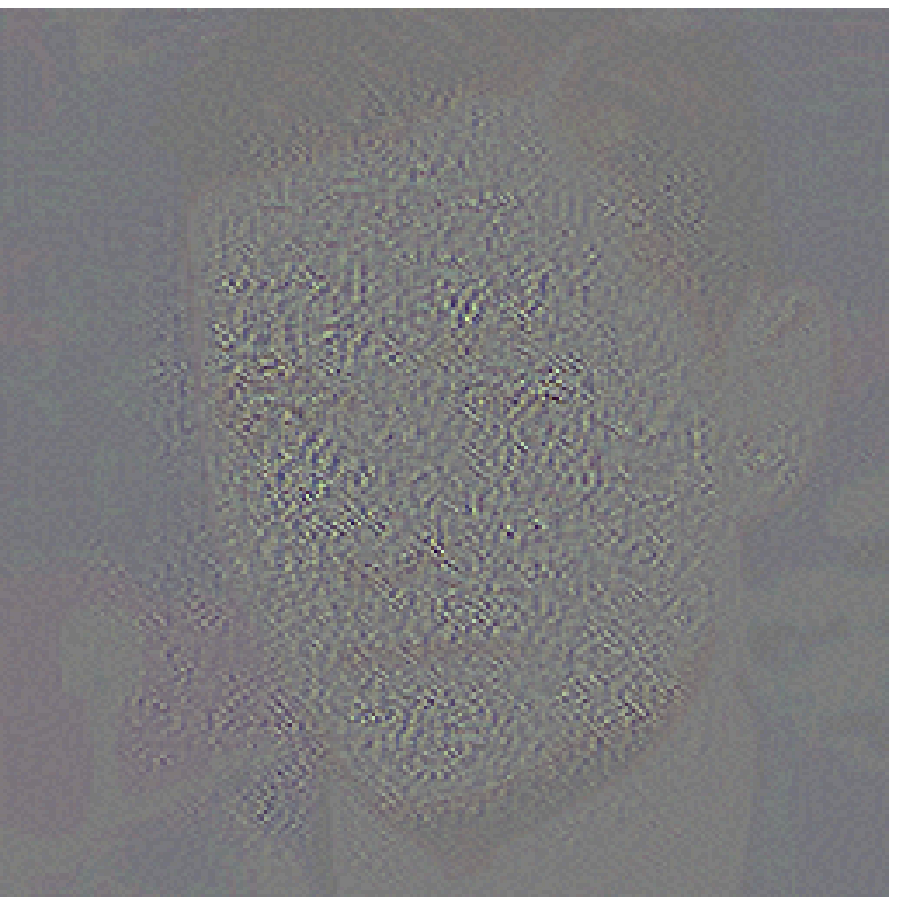}}
    \hspace{-0.08cm}
    \subfigure[]{\includegraphics[height=1.65cm]{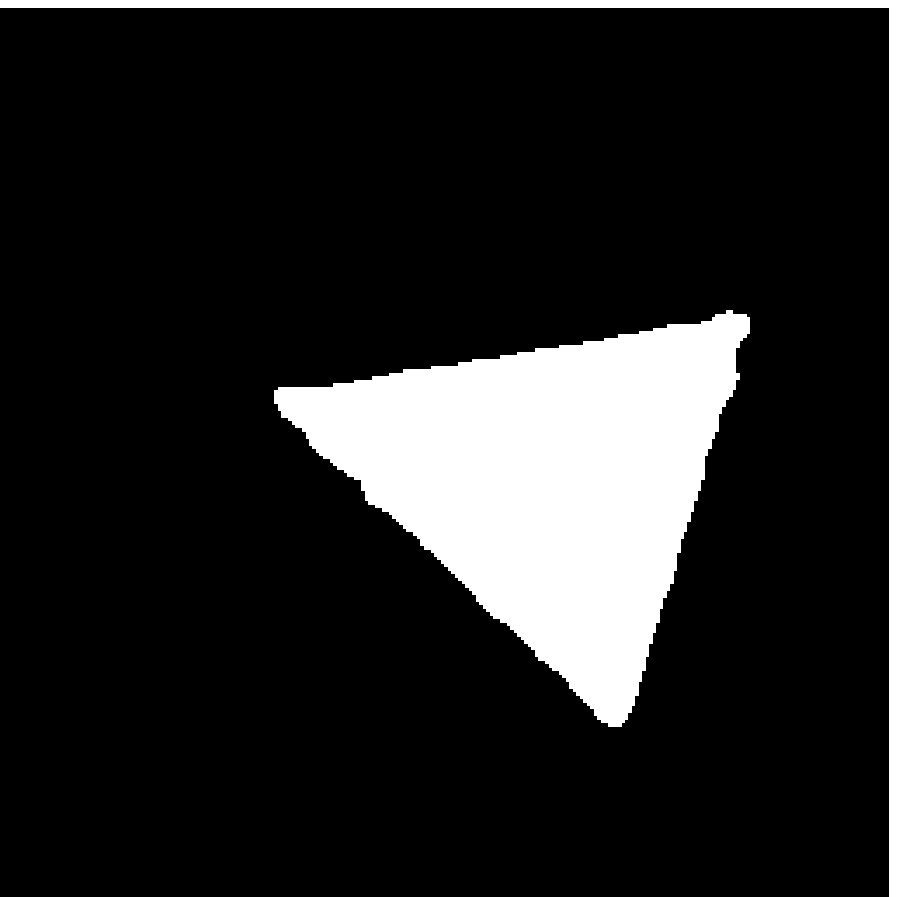}}\addtocounter{subfigure}{-4}



    \vspace{-0.4cm}

    \caption{Resulting examples (selected from: FaceSwap \& 500 iters). (a) Adversarial mask. (b) Adversarial images. (c) Adversarial perturbations (min-max scaled for display). (d) Segmentation outputs. Original images and their ground-truth masks are shown in Fig.\ref{fig:IAP_c}.}
    \label{fig:IAP_s}
\vspace{-0.5cm}
\end{figure}

 We also purposely fabricate incorrect segmentation outputs for fake images to mislead tampering localization judgement. In this \linebreak scenario, the adversarial ground-truth masks $m^{\text{adv}}_i$, as shown in Fig.\ref{fig:IAP_s} (a), are prepared in advance. The goal is to slightly adapt the pixels of $x_i$, making segmentation output as similar as possible to $m^{\text{adv}}_i$. To this end, we define a weighted objective function $\mathcal{L} = \mathcal{L}_{\text{act}}(x_i,l^{\text{adv}}_i) + \lambda \cdot \mathcal{L}_{\text{seg}}(x_i,m^{\text{adv}}_i)$, where $\lambda$ is set to 1.8 empirically. Then, this objective function is substituted into Eq.(\ref{g_IAP}) for iterative computations. Here, the adversarial label $l^{\text{adv}}_i$ is assigned a value close to but less than 0.5 (for example 0.45)\footnote{Recall that label `fake' takes the value 0 in this paper.} in this scenario. Such a setting loosens the interaction between classification and segmentation, while still ensuring that the zero zone $h_{0,i}$ is activated.

\par Figure \ref{fig:IAP_s} shows the corresponding resulting examples. Obviously, the fabricated masks (d) closely match the adversarial ones (a), and are much different from the original ones. For quantitative \linebreak evaluation, it is time-consuming to manually create a meaningful adversarial mask for every image. For a quick experiment, we assume that an adversarial mask follows a simple geometry, i.e., a triangle, as shown by the bottom image in Fig.\ref{fig:IAP_s} (a). The location and orientation of the triangle are randomly assigned for each image. The results of the quantitative evaluation are summarized in Table \ref{tab_IAP_s}. The IoU scores w.r.t the adversarial masks were as high as $48.52\%$ while those w.r.t the original masks were reduced on average from $43.21\%$ to $22.04\%$. This means that adding IAPs leads the segmentation outputs being more similar to the adversarial masks. It may be difficult to further reduce the IoU scores in this quick experiment because a triangle with a random location and orientation may partly overlap the original highlighted area. Interestingly, the classification accuracy got a rise in this scenario. This is because the setting $l^{\text{adv}}_i=0.45$ forced the zero zone to activate, which provided a chance to correct previously misclassified outputs.  In addition, fabricating a segmentation output needs 100 or more iterations, which has a higher computational cost than falsification for classification.

\begin{table}[t]
	\begin{center}

        \caption{Performance of IAPs (fabricating classification outputs). Only `fake' images were considered in this scenario, so scores in the `original' rows differ from those in Table \ref{tab_IAP_c}.}
        \vspace{-0.0cm}
        \begin{tabular}{c}
			
			\scalebox{0.23}[0.23]{\includegraphics{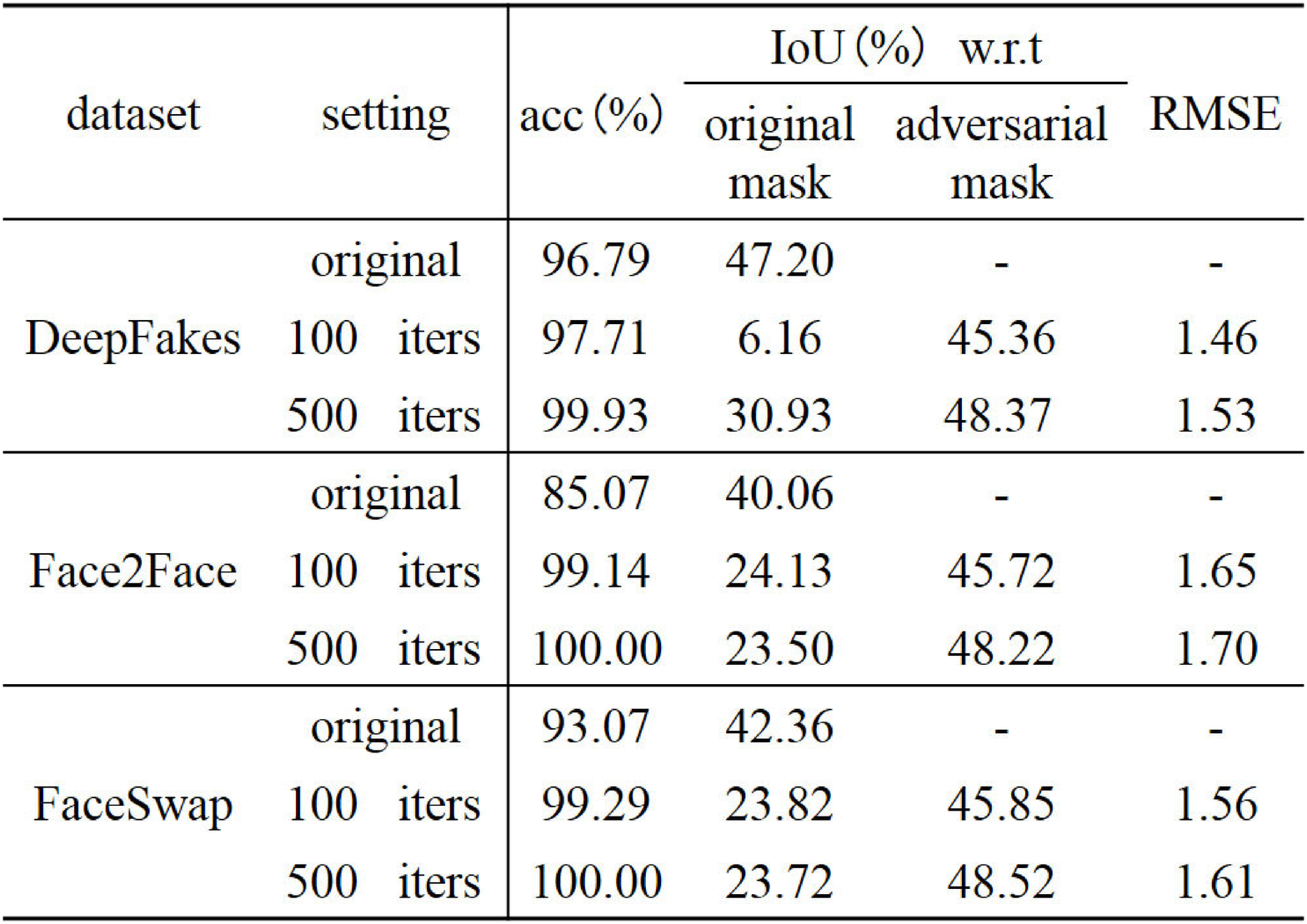}}
			
		\end{tabular}

		\label{tab_IAP_s}
	\end{center}
\vspace{-0.9cm}
\end{table}

\vspace{-0.08cm}
\subsection{Universal adversarial attacks}
\vspace{-0.08cm}
\label{s2}
In this scenario, we aim to craft a data-free UAP $\Xi$ that can flip the classification outputs for most images. The iterative procedure starts with a random perturbation in which the entries obey a uniform distribution $[-\epsilon,\epsilon]$. For the autoencoder network \cite{BTAS}, it is nontrivial to activate the latent neurons at the initial iterations due to the small magnitude of the perturbation-only input. To overcome this problem, we designed a new objective function:
\begin{equation}
\mathcal{L}(\Xi) = \exp\Big(\frac{\kappa\cdot a_{i,1-c}}{a_{i,c}}\Big)-a_{i,c},
\label{loss_UAP}
\end{equation}
\par\noindent where $a_{i,c}$ and $a_{i,1-c}$ denote the activation energy of $h_{i,c}$ and $h_{i,1-c}$, respectively. The iterative procedure for updating $\Xi$ can be formulated as
\begin{equation}
\Xi^{(n+1)} = \text{Clip}_{\epsilon}\big\{\Xi^{(n)}-\alpha(\nabla \mathcal{L}(\Xi^{(n)})) \big\}.
\label{g_UAP}
\end{equation}

\begin{figure}[t]
    \centering

    \subfigure{\includegraphics[height=4.2cm]{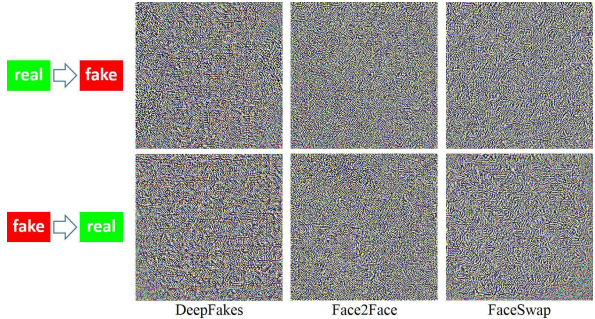}}
    \vspace{-0.4cm}

    \caption{UAPs (min-max scaled for display) generated for fabricating classification outputs.}
    \label{fig:UAP}
\vspace{-0.5cm}
\end{figure}

\begin{table}[t]
	\begin{center}

        \caption{Performance of UAPs: image-agnostic property and transferability across unseen datasets.}
        \vspace{-0.30cm}
        \begin{tabular}{c}
			
			\scalebox{0.211}[0.211]{\includegraphics{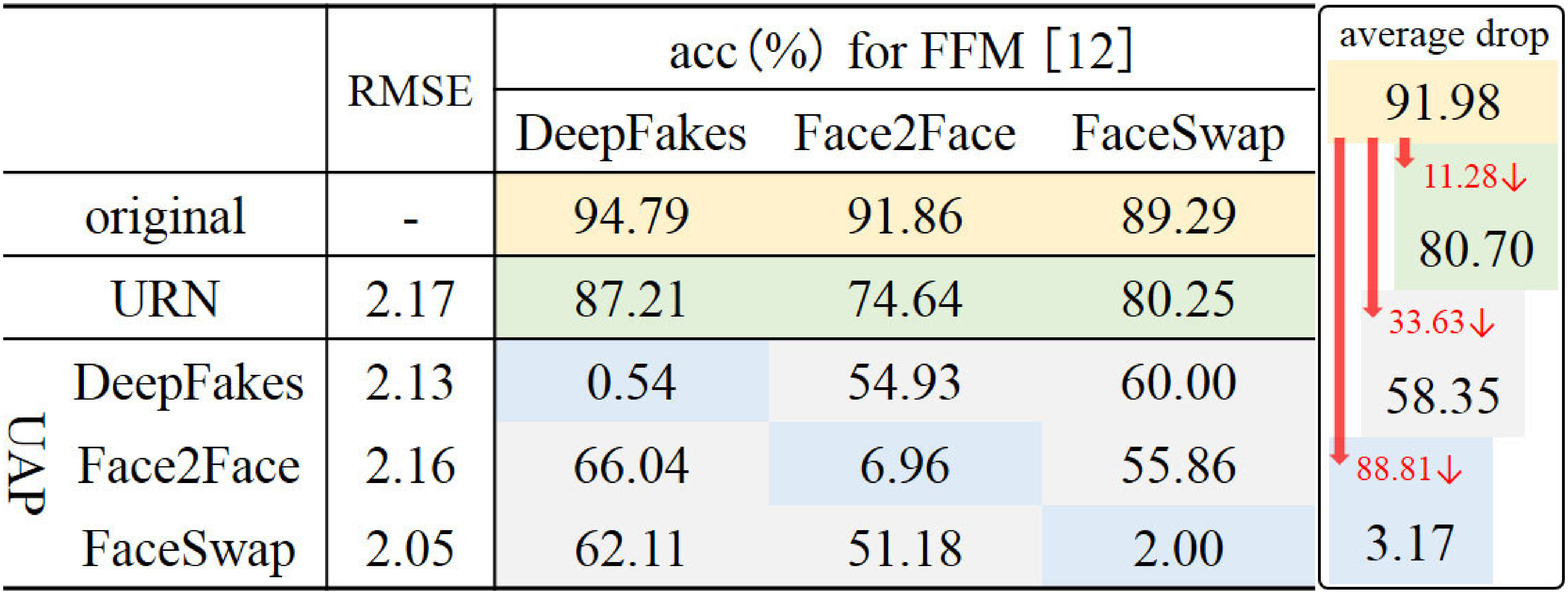}}
			
		\end{tabular}

		\label{tab_UAP_set}
	\end{center}
\vspace{-1.0cm}
\end{table}

\par\noindent 
Clearly, minimizing the objective function in Eq.(\ref{loss_UAP}) is equivalent to maximizing $a_{i,c}$. At the beginning of the iteration procedure, both $a_{i,c}$ and $a_{i,1-c}$ are small, and we have $a_{i,c} \approx a_{i,1-c}$ in most cases. If we set $\kappa>1.0$, the exponential term in Eq.(\ref{loss_UAP}) can amplify the difference between $a_{i,c}$ and $a_{i,1-c}$ so as to initially provide a sufficient loss. After several iterations, the value of the exponential term approaches 1 while the negative term becomes a potency to further minimize the loss.

\par Only 280 iterations were needed to generate an effective UAP in our experiments, which were performed in a data-free manner. This demonstrates that over-firing is a resource-conserving approach for attackers. Looking at the generated UAPs in Fig.\ref{fig:UAP}, we observe similar local textures organized with different styles. For each dataset, two UAPs which can over-fire the two disjoint zones, are crafted separately. 
As shown in Table \ref{tab_UAP_set}, classification accuracy was reduced on average from $91.98\%$ (orange shading) to $3.17\%$ (blue shading), which demonstrates the image-agnostic property of UAPs. Evaluation of transferability across unseen datasets showed that adding UAPs reduced accuracy $33.63\%$ on average (gray shading). As a reference, uniform random noises (URNs) with the same level of strength only achieved a slight drop of $11.28\%$ (green shading).

\begin{table}[t]
	\begin{center}

        \caption{Performance of UAPs: transferability across unseen FFMs.}
        \vspace{-0.30cm}
        \begin{tabular}{c}
			
			\scalebox{0.211}[0.211]{\includegraphics{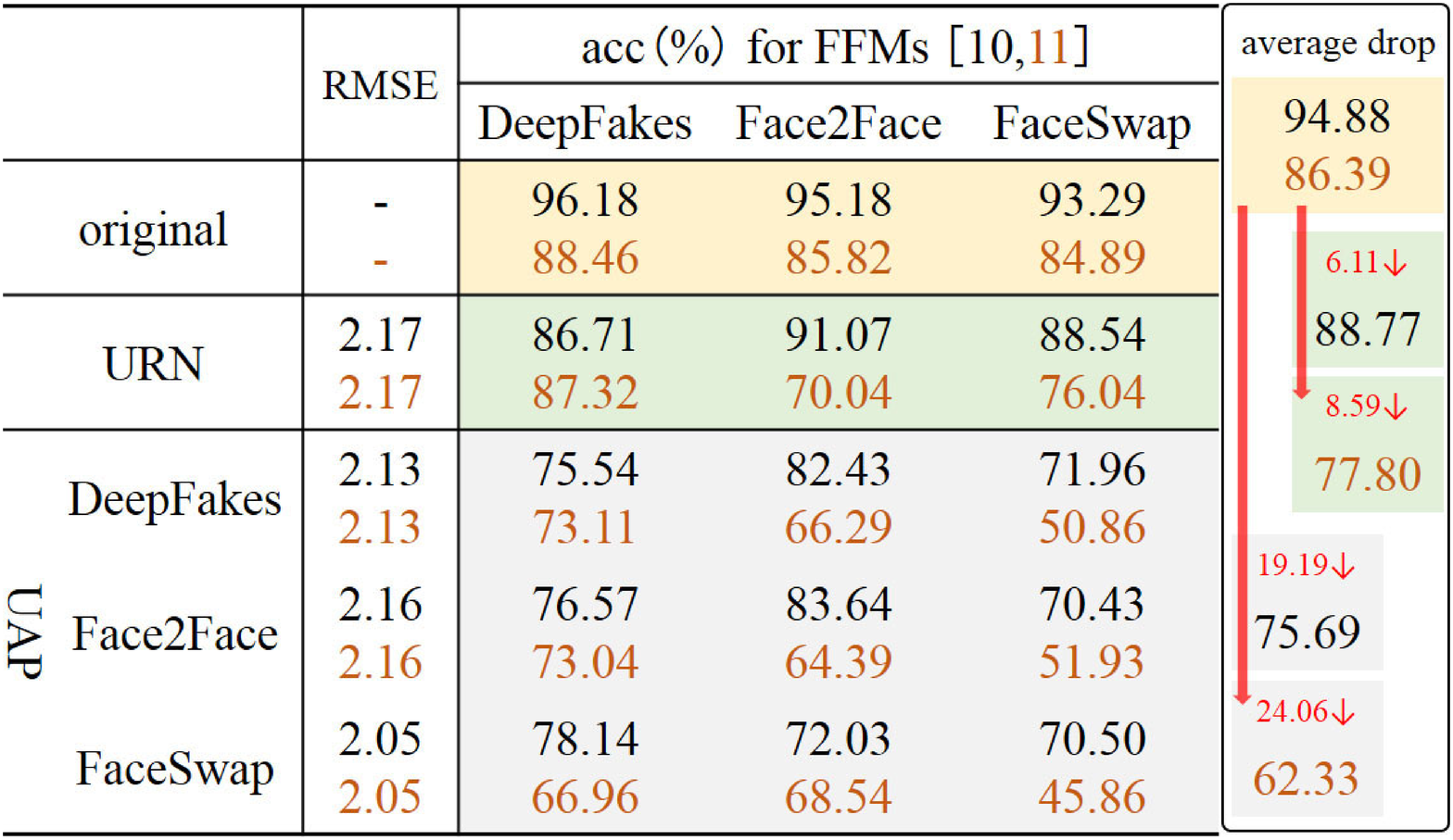}}
			
		\end{tabular}

		\label{tab_UAP_mod}
	\end{center}
\vspace{-0.9cm}
\end{table}

\par UAP transferability across unseen models is highly desirable, \linebreak especially when the adversary has no knowledge of such models. The UAPs computed for the FFM \cite{BTAS} directly transfer to attack other target FFMs \cite{MesoNet,FT}. Table \ref{tab_UAP_mod} shows the evaluation results (brown ink for \cite{FT}).  Note that the FFM \cite{FT} is reimplemented here without including the residual module. As shown in Table \ref{tab_UAP_mod}, the UAPs \linebreak reduced average accuracies by 19.19$\%$ and 24.06$\%$, respectively, for the two unseen FFMs. Again, the UAPs outperform the URNs, \linebreak which demonstrates that an adversarial attack based on over-firing can probe special noise patterns that contaminate the learned features even though the network architectures differ. Interestingly, if we retain the residual module for the FFM \cite{FT}, the UAPs and URNs have comparable performance, i.e., an average drop of around $30\%$ (not shown in Table \ref{tab_UAP_mod}). The reasons for this are twofold. First, the residual module extracts high-frequency details, which renders the whole model sensitive to both kinds of perturbations. Second, the residual maps may lie in a region far from the manifold of original data, so transferability is no longer effective.

\begin{table}[!]
	\begin{center}

        \caption{Results of subjective assessment.}
        \vspace{-0.30cm}
        \begin{tabular}{c}
			
			\scalebox{0.209}[0.209]{\includegraphics{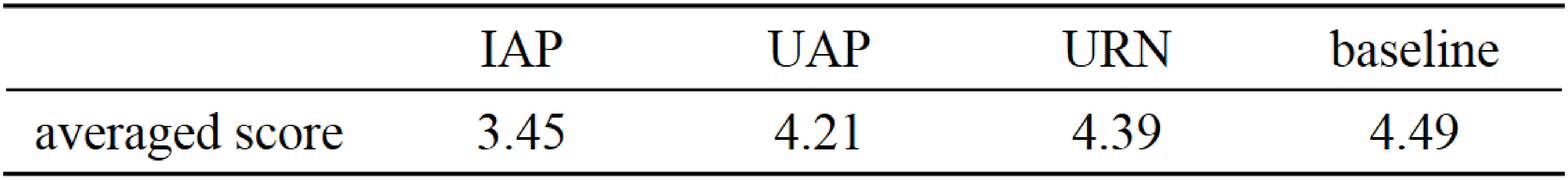}}
			
		\end{tabular}

		\label{tab_sa}
	\end{center}
\vspace{-0.9cm}
\end{table}

\vspace{-0.08cm}
\subsection{Subjective assessment}
\vspace{-0.08cm}
\label{sec:sa}
Imperceptibility is an essential requirement for adversarial perturbations. However, the objective criteria like RMSE can not dutifully reflect the perceptual loss caused by local but condensed distortions. We conducted subjective assessment for the visual quality of adversarial images by using the degradation category rating method \cite{ITU}. We asked evaluators to compare a clean image with its noisy version and then to rate the quality degradation on a 5-point mean opinion score scale, with 5 being the least degradation. We selected IAPs (all groups of `20 iters'), UAPs and URNs as the distortion sources, and compared the clean image with itself as a baseline. As such, there were four conditions, each consisting of 300 pairs selected at random. We exhibited some pairs at \url{https://nii-yamagishilab.github.io/Samples-Rong/Image-AE-attack/}.
\par The evaluation was performed through a web-based interface, with each web page horizontally displaying a pair of images. We grouped 21 pairs into a set, and each evaluator was permitted to \linebreak evaluate at most 5 sets. We collected a total of 39680 legitimate scores from 775 evaluators. As shown in Table \ref{tab_sa}, the UAP and URN scores were comparable and close to baseline 4.49. This demonstrates that UAPs are visually negligible. The lower IAP score suggests that local but condensed distortions are prone to serious perceptual losses. Therefore, future work on individual adversarial attacks should enforce a constraint based on local smoothness.

\vspace{-0.1cm}
\section{Conclusion}
\vspace{-0.1cm}
Our study had demonstrated the existence of adversarial perturbations for FFMs. The gradient-based iterative procedure generates IAPs that can be used to fabricate classification and segmentation outputs. Without the need for training data, UAPs are crafted to over-fire the latent neurons. The objective function we designed provides a sufficient initial loss, which makes the over-firing practical with lightweight computational cost. Our experiments demonstrated that UAPs have transferability across unseen datasets and unseen FFMs, and our subjective assessment showed that the distortions incurred by UAPs are visually negligible. This work can serve as a baseline for evaluating the adversarial security of FFMs. Future work includes developing a defense module that can protect FFMs against adversarial attacks.





\begin{thebibliography}{00}

\bibitem{df} https://github.com/deepfakes/faceswap/

\bibitem{fs} https://github.com/MarekKowalski/FaceSwap/

\bibitem{lip} P. Korshunov, S. Marcel, ``Speaker Inconsistency Detection in Tampered Video,'' in \emph{Proc. $26^{th}$ Eur. Signal Process. Conf. (EUSIPCO)}, pp.2375-2379, 2018.

\bibitem{eye} Y. Li, M.C. Chang, S.W. Lyu, ``In Ictu Oculi: Exposing AI Generated Fake Face Videos by Detecting Eye Blinking,'' in \emph{Proc. IEEE Int. Workshop Inform. Forensics Secur. (WIFS)}, pp.1-7, 2018.

\bibitem{head} X. Yang, Y.Z. Li, S.W. Lyu, ``Exposing Deep Fakes Using Inconsistent Head Poses,'' in \emph{Proc. $44^{th}$ IEEE Int. Conf. Acoust., Speech, Signal Process. (ICASSP)}, pp.8261-8265, 2019.

\bibitem{iqm} D. Wen, H. Han, A.K. Jain, ``Face Spoof Detection with Image Distortion Analysis,'' \emph{IEEE Trans. Inform. Forensics Secur.}, 10(4):746-761, 2015.

\bibitem{color} H.D. Li, B. Li, S.Q. Tan, J.W. Huang, ``Detection of Deep Network Generated Images Using Disparities in Color Components,'' \emph{arXiv preprint:1808.07276v2}, 2019.

\bibitem{warping} Y.Z. Li, S.W. Lyu, ``Exposing DeepFake Videos by Detecting Face Warping Artifacts,'' in \emph{Proc. 2019 IEEE Conf. Comput. Vis. Pattern Recogn. Workshop (CVPRW)}, pp.46-52, 2019.

\bibitem{Mo} H.X. Mo, B.L. Chen, W.Q. Luo, ``Fake Faces Identification via Convolutional Neural Network,'' in \emph{Proc. $6^{th}$ ACM Workshop Inform. Hiding \& Multimedia Secur. (IH\&MMSec)}, pp.43-47, 2018.

\bibitem{MesoNet} D. Afchar, V. Nozick, J. Yamagishi, I. Echizen, ``MesoNet: A Compact Facial Video Forgery Detection ,'' in \emph{Proc. IEEE Int. Workshop Inform. Forensics Secur. (WIFS)}, pp.1-7, 2018.


\bibitem{FT} D. Cozzolino, J. Thies, A. R{\"o}ssler, C. Riess, M. Nie{\ss}ner, L. Verdoliva, ``ForensicTransfer: Weakly-Supervised Domain Adaptation for Forgery Detection,'' \emph{arXiv preprint:1812.02510v1}, 2018.

\bibitem{BTAS} H.H. Nguyen, F.M. Fang, J. Yamagishi, I. Echizen, ``Multi-task Learning for Detecting and Segmenting Manipulated Facial Images and Videos,'' \emph{Proc. $10^{th}$ IEEE Int. Conf. Biom. Theory, Appl. Syst. (BTAS)}, pp.1-8, 2019.

\bibitem{survey1} N. Akhtar, A. Mian, ``Threat of Adversarial Attacks on Deep Learning in Computer Vision: A Survey,'' \emph{IEEE Access}, 6:14410-14430, 2018.

\bibitem{survey2} X.Y. Yuan, P. He, Q.L. Zhu, X.L. Li, ``Adversarial Examples: Attacks and Defenses for Deep Learning,'' \emph{IEEE Trans. Neural Net. Learn. Syst.}, 30(9):2805-2824, 2019.

\bibitem{face_attack} M. Sharif, S. Bhagavatula, L. Bauer, M. K. Reiter, ``Accessorize to a Crime: Real and Stealthy Attacks on State-of-the-Art Face Recognition,'' \emph{Proc. ACM SIGSAC Conf. Comput. Commun. Secur.}, pp.1528-1540, 2016.

\bibitem{obj_attack} C.H. Xie, J.Y. Wang, Z.H. Zhang, Y.Y. Zhou, L.X. Xie, A. Yuille, ``Adversarial Examples for Semantic Segmentation and Object Detection,'' \emph{Proc. Int. Conf. Comput. Vis. (ICCV)}, pp.1378-1387, 2017.

\bibitem{ss_attack} J.H. Metzen, M. C. Kumar, T. Brox, V. Fischer, ``Universal Adversarial Perturbations Against Semantic Image Segmentation,'' \emph{Proc. IEEE Conf. Comput. Vis. Pattern Recognit. (CVPR)}, pp.2755-2764, 2017.

\bibitem{nlp_attack} R. Jia, P. Liang, ``Adversarial Examples for Evaluating Reading Comprehension Systems,'' \emph{Proc. Conf. Empirical Methods Natural Lang. Process. (EMNLP)}, pp.2021-2031, 2017.

\bibitem{md_attack} K. Grosse, N. Papernot, P. Manoharan, M. Backes, P. McDaniel, ``Adversarial Examples for Malware Detection,'' \emph{Proc. Eur. Symp. Res. Comput. Secur. (ESORICS)}, pp.62-79, 2017.

\bibitem{ITU} ITU-R Recommendation BT.1866, ``Objective Perceptual Video Quality Measurement Techniques for Broadcasting Applications Using Low Definition Television in the Presence of a Full Reference Signal,'' 2010.

\bibitem{data} A. R{\"o}ssler, D. Cozzolino, L. Verdoliva, C. Riess, J. Thies, M. Nie{\ss}ner, ``FaceForensic++: Learning to Detect Manipulated Facial Images,'' \emph{arXiv preprint:1901.08971v3}, 2019.

\bibitem{BFGS} C. Szegedy, W. Zaremba, I. Sutskever, J. Bruna, D. Erhan, I. Goodfellow, R. Fergus, ``Intriguing Properties of Neural Networks,'' \emph{arXiv preprint:1312.6199v4}, 2014.

\bibitem{FGSM} I. Goodfellow, J. Shlens, C. Szegedy, ``Explaining and Harnessing Adversarial Examples,'' \emph{arXiv preprint:1412.6572v3}, 2015.

\bibitem{BIM} A. Kurakin, I. Goodfellow, S. Bengio, ``Adversarial Examples in the Physical World,'' \emph{arXiv preprint:1607.02533v4}, 2017.

\bibitem{Deepfool} S. Moosavi-Dezfooli, A. Fawzi, P. Frossard, ``DeepFool: A Simple and Accurate Method to Fool Deep Neural Networks,'' \emph{Proc. 2016 IEEE Conf. Comput. Vis. Pattern Recogn. (CVPR)}, pp.2574-2582, 2016.

\bibitem{UAP} S. Moosavi-Dezfooli, A. Fawzi, O. Fawzi, P. Frossard, \linebreak``Universal Adversarial Perturbations,'' \emph{Proc. 2017 IEEE Conf. Comput. Vis. Pattern Recogn. (CVPR)}, pp.86-94, 2017.

\bibitem{overfire} K. Mopuri, A. Ganeshan, R. Babu, ``Generalizable Data-Free Objective for Crafting Universal Adversarial Perturbations,'' \emph{IEEE Trans. Pattern Anal. Mach. Intell.}, 41(10):2452-2465, 2019.

\end{thebibliography}

\end{document}